\documentclass[letterpaper, 10 pt, conference]{ieeeconf}  

\IEEEoverridecommandlockouts                              

\usepackage{graphics} 
\usepackage{svg} 
\usepackage{times} 
\usepackage{amsmath} 
\usepackage{amssymb}  
\usepackage{booktabs}

\usepackage{mathtools}
\usepackage{tabularx}
\usepackage{array}
\usepackage{makecell}
\usepackage{multirow}
\usepackage{multicol}
\usepackage{graphicx}
    \graphicspath{{figure/}}
\usepackage[font={small}]{subcaption}
\usepackage[font={small}]{caption}
\usepackage{xcolor}

\usepackage{listings}
\definecolor{codegreen}{rgb}{0,0.6,0}
\definecolor{codegray}{rgb}{0.5,0.5,0.5}
\definecolor{codepurple}{rgb}{0.58,0,0.82}
\definecolor{backcolour}{rgb}{0.95,0.95,0.92}

\lstdefinestyle{mystyle}{
  backgroundcolor=\color{backcolour},   commentstyle=\color{codegreen},
  keywordstyle=\color{magenta},
  numberstyle=\tiny\color{codegray},
  stringstyle=\color{codepurple},
  basicstyle=\ttfamily\scriptsize,
  breakatwhitespace=false,         
  breaklines=true,                 
  captionpos=t,                    
  keepspaces=true,        
  frame=single,
  showspaces=false,                
  showstringspaces=false,
  showtabs=false,                  
  tabsize=2
}

\usepackage{algorithm}
\usepackage{algpseudocode}

\newtheorem{definition}{Definition}

\newtheorem{problem}{Problem}

\makeatletter
\let\NAT@parse\undefined
\makeatother
\usepackage{url}
\definecolor{mycitecolor}{RGB}{71, 191, 38}
\definecolor{mylinkcolor}{RGB}{40, 115, 201}
\definecolor{LightGray}{gray}{0.9}

\usepackage{hyperref}
\hypersetup{
  colorlinks=true,
  citecolor=mycitecolor,
  linkcolor=mylinkcolor,
  urlcolor=mycitecolor,
}

\title{\LARGE \bf
CurricuLLM: Automatic Task Curricula Design for Learning Complex Robot Skills using Large Language Models
}

\author{Kanghyun Ryu$^{1}$, Qiayuan Liao$^{1}$, Zhongyu Li$^{1}$, Payam Delgosha$^{2}$, Koushil Sreenath$^{1}$, Negar Mehr$^{1}$
\thanks{*This work is supported by the National Science Foundation, under grants ECCS-2438314 CAREER Award, CNS-2423130, and CCF-2423131}
\thanks{$^{1}$Mechanical Engineering,
        University of California Berkeley
        {\tt\small \{kanghyun.ryu, qiayuanl, zhongyu\_li, koushils, negar\}@berkeley.edu}}%
\thanks{$^{2}$This work was done when Payam was a research assistant professor at Computer Science department, University of Illinois at Urbana-Champaign. Payam is currently at Apple. {\tt\small delgosha@illinois.edu}}
}

\begin{document}

\maketitle
\thispagestyle{empty}
\pagestyle{empty}

\begin{abstract}

Curriculum learning is a training mechanism in reinforcement learning (RL) that facilitates the achievement of complex policies by progressively increasing the task difficulty during training. However, designing effective curricula for a specific task often requires extensive domain knowledge and human intervention, which limits its applicability across various domains. 
Our core idea is that large language models (LLMs), with their extensive training on diverse language data and ability to encapsulate world knowledge, present significant potential for efficiently breaking down tasks and decomposing skills across various robotics environments. Additionally, the demonstrated success of LLMs in translating natural language into executable code for RL agents strengthens their role in generating task curricula. 
In this work, we propose CurricuLLM, which leverages the high-level planning and programming capabilities of LLMs for curriculum design, thereby enhancing the efficient learning of complex target tasks. CurricuLLM consists of: (Step 1) Generating sequence of subtasks that aid target task learning in natural language form, (Step 2) Translating natural language description of subtasks in executable task code, including the reward code and goal distribution code, and (Step 3) Evaluating trained policies based on trajectory rollout and subtask description. We evaluate CurricuLLM in various robotics simulation environments, ranging from manipulation, navigation, and locomotion, to show that CurricuLLM can aid learning complex robot control tasks. In addition, we validate humanoid locomotion policy learned through CurricuLLM in real-world. Project website is \url{https://iconlab.negarmehr.com/CurricuLLM/}

\end{abstract}

\section{INTRODUCTION}

Deep reinforcement learning (RL) has achieved notable success across various robotics tasks, including manipulation~\cite{gu2017deep}, navigation~\cite{zhu2017target}, and locomotion~\cite{li2021reinforcement}. However, RL requires informative samples for learning, and obtaining these from a random policy is highly sample-inefficient, especially for complex tasks. In contrast, human learning strategies differ significantly from random action trials; they typically start with simpler tasks and progressively increase difficulty. Curriculum learning, inspired by this structured approach of learning, aims to train models in a meaningful sequence~\cite{soviany2022curriculum}, gradually enhancing the complexity of the training data~\cite{pentina2015curriculum} or the tasks themselves~\cite{narvekar2020curriculum}. Particularly in RL, curriculum learning improves training efficiency by focusing on simpler tasks that can provide informative experiences to reach more complex target task, instead of starting from scratch~\cite{narvekar2020curriculum}.

\begin{figure}
    \centering
    \includegraphics[width=\linewidth]{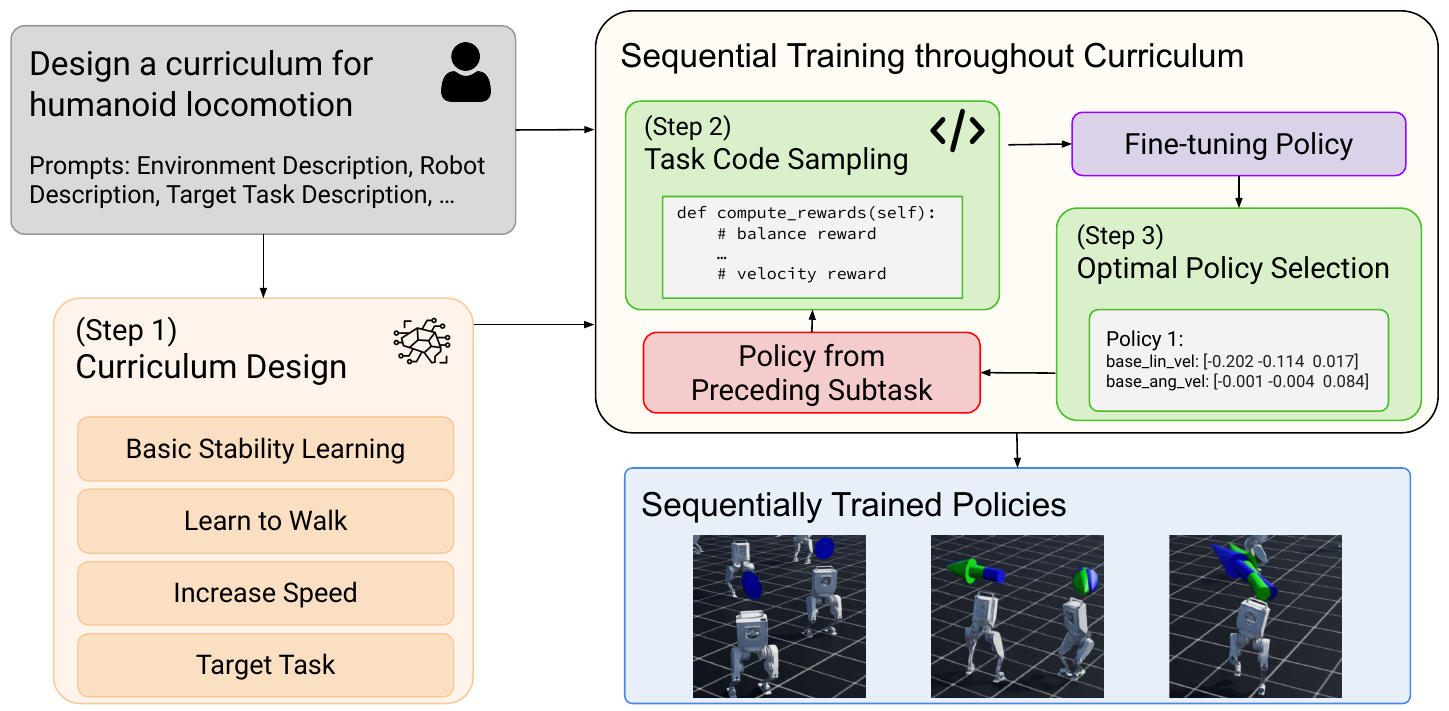}
    \caption{CurricuLLM takes natural language description of environments, robots, and target task that we wish the robot to learn, and then generates a sequence of subtasks. In each subtasks, it samples different task codes and evaluates the resulting trained policy to find the policy which is best aligned within the current subtask. These iterations are repeated throughout the curriculum subtasks to sequentially train a policy that reaches complex target task.}
    \label{fig:main}
\end{figure}

Although effective, designing a good curriculum is challenging. Manual curriculum design often necessitates the costly intervention of human experts~\cite{li2024reinforcement, tang2021learning, tidd2020guided} and is typically restricted to a limited set of predefined tasks~\cite{rudin2022learning}. Consequently, several works focused on automatic curriculum learning (ACL). To generate task curricula, ACL requires the ability of both determining subtasks aligned with the target task, ranking the difficulty of each subtask, and organizing them in ascending order of difficulty~\cite{wang2021survey}. However, autonomously evaluating the relevance and difficulty of these subtasks remains unresolved. As a result, ACL has been limited to initial state curricula~\cite{florensa2017reverse, ivanovic2019barc}, goal state curricula~\cite{forestier2022intrinsically}, or environment curricula~\cite{rudin2022learning, wang2023curriculum}, rather than task-level curricula. 

Meanwhile, in recent years, large language models (LLMs) trained on extensive collections of language data~\cite{achiam2023gpt, touvron2023llama, reid2024gemini} have been recognized as repositories of world knowledge expressed in linguistic form~\cite{wei2022chain}. Leveraging this world knowledge, LLMs have demonstrated their capabilities in task planning~\cite{brohan2023can} and skill decomposition for complex robotic tasks~\cite{wang2023voyager, wang2024gensim}. Furthermore, the programming skills of LLMs enabled smooth integration between high-level language description and robotics through API call composition~\cite{liang2023code, singh2023progprompt}, simulation environment generation~\cite{wang2024robogen, liang2024environment}, or reward design~\cite{ma2024eureka, li2024auto}.

In this paper, we introduce CurricuLLM, which leverages the reasoning and coding capabilities of LLMs to design curricula for complex robotic control tasks. Our goal is to autonomously generate a series of subtasks that facilitate the learning of complex target tasks without the need for extensive human intervention. Utilizing the LLM's task decomposition and coding, CurricuLLM autonomously generates sequences of subtasks along with appropriate reward functions and goal distributions for each subtask, enhancing the efficiency of training complex robotic policies. 

Our contribution can be summarized in threefold. First, we propose CurricuLLM, a task-level curriculum designer that uses the high-level planning and code writing capabilities of LLMs. Second, we evaluate CurricuLLM in diverse robotics simulation environments ranging from manipulation, navigation, and locomotion, demonstrating its efficacy in learning complex control tasks. Finally, we validate the policy trained with CurricuLLM on the Berkeley Humanoid~\cite{liao2024berkeley}, illustrating that the policy learned through CurricuLLM can be transferred to the real world.

\section{Related Works}

\subsection{Curriculum Learning}

 In RL, curriculum learning is recognized for enhancing sample efficiency~\cite{schaul2015prioritized}, addressing previously infeasible challenging tasks~\cite{portelas2021automatic}, and facilitating multitask policy learning~\cite{jabri2019unsupervised}. Key elements of curriculum learning include the difficulty measure, which ranks the difficulty of each subtask, and training scheduling, which arranges subtasks at an appropriate pace~\cite{wang2021survey}. The teacher-student framework, for example, has a teacher agent that monitors the progress of the student agent, recommending suitable tasks or demonstrations accordingly~\cite{matiisen2019teacher, clark2024adaptive}. However, this method requires a predefined set of tasks provided by human experts or a teacher who has superior knowledge of the environment. Although self-play has been proposed as a means to escalate opponent difficulty~\cite{silver2018general, du2022takes}, it is limited to competitive multi-agent settings and may converge to a local minimum.

 An appropriate difficulty measure is also crucial in curriculum learning. In goal-conditioned environments, it is often suggested to start training from a goal distribution close to the initial state~\cite{forestier2022intrinsically, lee2023cqm} or an initial state distribution that is in proximity to the goal state~\cite{florensa2017reverse, ivanovic2019barc} to regulate difficulty. However, these methods are limited to goal-conditioned environment where the task difficulty is correlated with how ``far'' the goal is from the start location. In this work, we use LLM to provide a more general method to measure task difficulty and design curricula.

The most closely related work to CurricuLLM is DrEureka~\cite{ma2024dreureka} and Eurekaverse~\cite{liang2024environment}, which utilizes LLMs to generate an domain randomization parameters, such as gravity or mass, or environment curriculum, such as terrain height. Especially, Eurekaverse employs a co-evolution mechanism that gradually increases the complexity of the environment by using the LLM for environment code generation. Compared to Eurekaverse, our method focuses on the \emph{task} curriculum, which focuses on task break-down for learning complex robotic tasks compared to the Eurekaverse that focuses on generalization across different environments. 

\begin{figure*}
    \centering
    \includegraphics[width=0.9\textwidth]{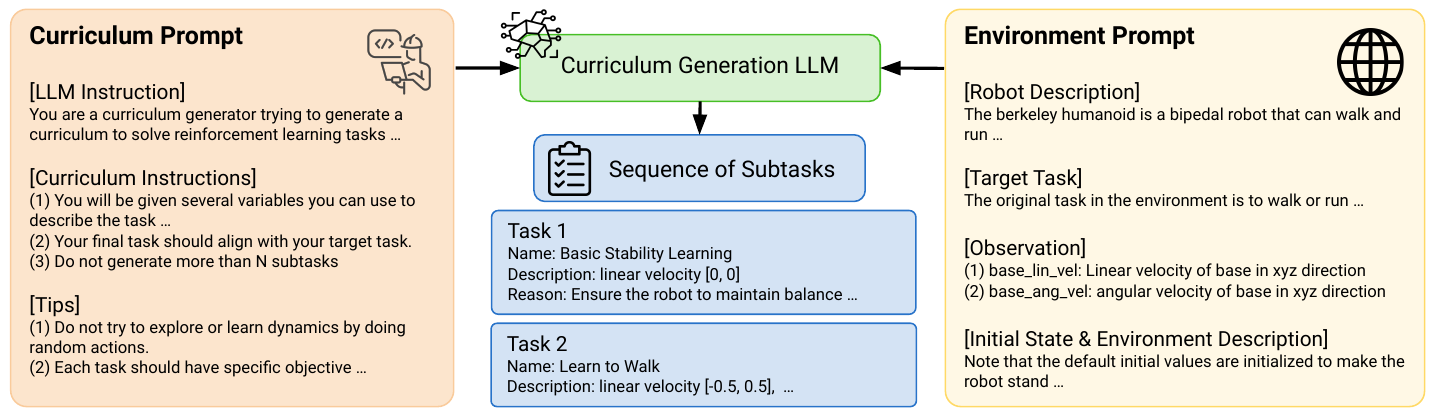}
    \caption{Curriculum generation LLM receives the natural language form of a curriculum prompt as well as the environment description to generate a sequence of subtasks. Our prompt includes instruction for tje curriculum designer, rules for how to describe the subtasks, and other tips on describing the curriculum. Environment description consists of the robot and its state variable description, the target task, and the initial state description.}
    \label{fig:curriculum generation}
\end{figure*}

\subsection{Large Language Model for Robotics}

\noindent \textbf{Task Planning.} The robotics community has recently been exploring the use of LLMs for high-level task planning~\cite{huang2022language, singh2023progprompt}. However, these methods are limited to task planning within predefined finite skill sets and suffers when LLMs' plan is not executable within given skill sets or environment~\cite{brohan2023can, singh2023progprompt, shah2022value}. In contrast, Voyager~\cite{wang2023voyager} introduces an automatic skill discovery, attempting to learn new skills that is not currently available but required for open-ended exploration. Nonetheless, skills in Voyager are limited to composing discrete actions and is inapplicable to continuous control problems. In this work, we propose the automatic generation of task curriculum consisting of a sequence of subtasks that facilitates the efficient learning of robotic control tasks. To manage the control of robots with high degrees of freedom, we utilize the coding capabilities of LLMs to generate a reward function for each subtask and sequentially train each subtasks in given order.

\noindent \textbf{Reward Design.}
In continuation from works using natural language as a reward~\cite{mirchandani2021ella, goyal2019using}, several works have proposed using LLMs as a tool to translate language to reward. For example, \cite{yu2023language} uses LLMs to translate motion description to cost parameters, which are optimized using model predictive control (MPC). However, they are limited to changing the parameters in cost functions that are hand-coded by human experts. On the other hand, some works proposed directly using LLM~\cite{kwon2023reward} or vision-language model (VLM)~\cite{rocamonde2024visionlanguage} as a reward function, which observes agent behavior and outputs a reward signal. However, these approach require expensive LLM or VLM interaction during training. Most similar approaches with our work are~\cite{ma2024eureka, song2023self}, which leverage LLMs to generate reward functions and utilize a evolutionary search to identify the most effective reward function. However, these methods require an evaluation metric in their feedback loops, and their reward search tends to optimize specifically for this metric. Therefore, for tasks only described in natural language, as subtasks in our curriculum, finding reward function without these evaluation metric can be challenging. Additionally, their evolutionary search is highly sample-inefficient, contradicting the efficient learning objectives of curriculum learning. In contrast, our work divides a single complex task into a series of subtasks, and then employ a reasoning approach analogous to the chain of thoughts~\cite{wei2022chain} to generate reward functions for complex target tasks.



\section{Problem Formulation}
\label{sec:problem formulation}

    In this work, we consider task curriculum generation for learning control policies for complex robot tasks. First, we model a (sub)task as a goal-conditioned Markov Decision Process (MDP), formally represented by a tuple $m= (\mathcal{S}, \mathcal{G}, \mathcal{A}, p, r, \rho_g)$. Here, $\mathcal{S}$ is set of states, $\mathcal{A}$ is set of action, $p(s'|s,a)$ is a transition probability function, $r(s,a,s',g)$ is a reward function, $\mathcal{G}$ is a goal space, and $\rho_g$ is a goal distribution. We use subscript $n$ to describe $n_{\text{th}}$ subtask in our curriculum. Then, following~\cite{narvekar2020curriculum}, we formally define a task curriculum as:
    
    \begin{definition}[Task-level Sequence Curriculum.~\cite{narvekar2020curriculum}]
        A task-level sequence curriculum can be represented as an ordered list of tasks $C = [m_1, m_2, \dots, m_N]$ where if $i \leq j$ for some $m_i, m_j \in C$, then the task $m_i$ should be learned before task $m_j$.
    \end{definition}

    In our work, we aim to generate a task curriculum $C = [m_1, m_2, \dots, m_N]$ which helps learning a policy $\pi$ that maximizes the cumulative reward $V_{m_T}(\pi) = \mathbb{E}_{\pi} \big[ \sum_{t=0}^H \gamma^t r(s_t, a_t,s_{t+1}, g) \big]$ associated with target task $m_T$. 

    \begin{problem}
        For target task $m_T =  (\mathcal{S}, \mathcal{G}, \mathcal{A}, p, r_T, \rho_{g,T})$ and task curriculum $C = [m_1, m_2, \dots, m_N]$, we denote a policy trained with curriculum $C$ as $\pi_C$. Our objective for curriculum design is finding a curriculum $C$ that maximizes the reward on the target task $\arg \max_C V_{m_T}(\pi_C)$.
    \end{problem}

     We assume the state space $\mathcal{S}$, the action space $\mathcal{A}$, the transition probability function $p$, and the goal space $\mathcal{G}$ are fixed, i.e., they do not change between subtasks. Therefore, we can specify the task curriculum with the sequence of reward functions and goal distributions that are associated with the subtasks, $C_{(r, \rho_g)} = [(r_1, \rho_{g, 1}), (r_2, \rho_{g,2}), \dots, (r_N, \rho_{g,N})]$. Here, we express the reward function and goal distribution tuple $(r, \rho_g)$ with a programming code for simulation environment and define it as a \textit{task code}. Therefore, CurricuLLM's objective reduces to generating sequence of task codes $C_{(r, \rho_g)}$ that maximize the target task performance.



\section{Method}
\label{sec:method}

    Even for LLMs encapsulating world knowledge, directly generating the sequence of task codes $C_{(r, \rho_g)}$ can be challenging. Since LLM is known to show better reasoning capability by following step-by-step instructions~\cite{wei2022chain}, we divide our curriculum generation into three main modules (see Figure~\ref{fig:main}):
    \begin{itemize}
        \item Curriculum Design (Step 1): A curriculum generation LLM receives natural language descriptions of the robot, environment, and target task to generate sequences of language descriptions $C_l = [l_1, l_2, \dots, l_N]$ of the task sequence curriculum $C$. 
        \item Task Code Sampling (Step 2): A task code generation LLM generates $K$ task code candidates $(r_n^k, \rho_{g, n}^k), \, k=\{1,2,\dots,K\}$ for the given subtask description $l_n$. These are in the form of executable code and are used to fine-tune the policy trained for the previous subtask.
        \item Optimal Policy Selection (Step 3): An evaluation LLM evaluates policies $\pi_{n}^k,  \, k=\{1,2,\dots,K\}$ trained with different task code candidates $(r_n^k, \rho_{g, n}^k), \, k=\{1,2,\dots,K\}$ to identify the policy that best aligns with the current subtask. Selected policy $\pi_n^*$ is used as a pretrained policy for the next subtask.
    \end{itemize}

\subsection{Generating Sequence of Language Description}
\label{subsec:sequence of language}

    Leveraging the high-level task planning from LLMs, we initially ask an LLM to generate a series of language descriptions $C_l = [l_1, l_2, \dots, l_N]$ for a task-level sequence curriculum $C = [m_1, m_2, \dots, m_N]$ facilitating the learning of a target task $m_T$. Initially, the LLM is provided with a language description of the target task $l_T$ and a language description of the environmental information $l_E$ to generate an environment-specific curriculum. When generating curriculum, we query LLM to use the target task $m_T$ as a final task in curriculum $m_N$. Moreover, we require the LLM to describe the subtask using available state variables. This enables the LLM to generate curricula that are grounded in the environment information and ensures the generation of a reliable reward function later (discussed in Section~\ref{subsec:task code}). For example, to generate a curriculum for a humanoid to learn running, we query the LLM to generate a curriculum for following a velocity and heading angle command $\{(v_x, v_y, \theta): -2 \leq v_x, v_y \leq 2, -\pi \leq \theta \leq \pi \}$, while providing state variables, such as \textit{base linear velocity}, \textit{base angular velocity}, or \textit{joint angle}. Then, the LLM generates the sequence of subtasks descriptions such as (1) Basic Stability Learning: Maintain stability by minimizing the joint deviation and height deviation, (2) Learn to Walk: Follow low speed commands of the form $\{(v_x, v_y, \theta): -1 \leq v_x, v_y \leq 1, -\pi/2 \leq \theta \leq \pi/2 \}$, and others.

\begin{figure*}
    \centering
    \includegraphics[width=0.9\textwidth]{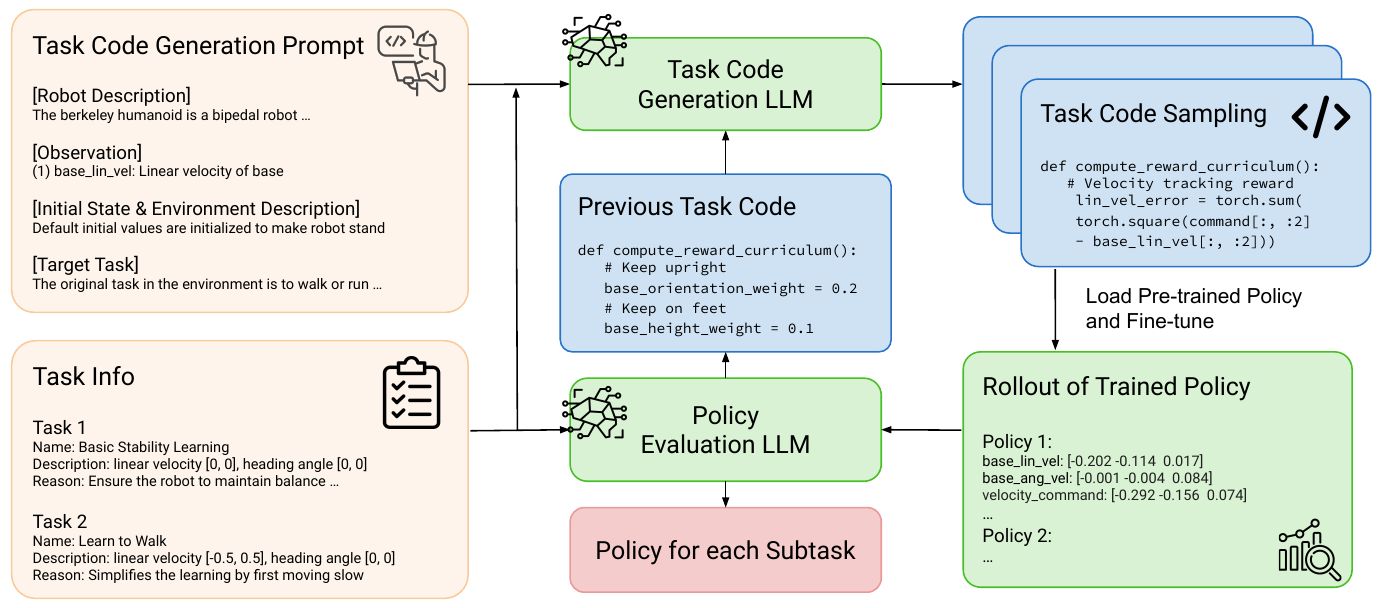}
    \caption{Our task code generation and evaluation framework in each subtask. Task code generation LLM takes the environment and target task description, current and past task information, and the reward function used for previous task. Then, $K$ task code candidates for current subtask is sampled and used for fine-tuning policies from previous subtask. Then, evaluation LLM receives the statistics of trajectory rollout from trained policy and find a policy that best aligns with current subtask description.}
    \label{fig:feedback loop}
\end{figure*}

\subsection{Task Code Generation}
\label{subsec:task code}

    After generating a curriculum with a series of language descriptions $C_l = [l_1, l_2, \dots, l_N]$, we should translate these language descriptions to a sequence of task codes $C_{(r, \rho_g)} = [(r_1, \rho_{g,1}), (r_2, \rho_{g,2}), \dots, (r_N, \rho_{g,N})]$. These task codes are described in executable code so that the RL policy can be trained on these subtasks. In the $n_{\text{th}}$ subtask which we denote by $m_n$, the task code generation LLM samples $K$ different task codes $(r_n^k, \rho_{g,n}^k), \, k \in \{1,2,\dots,K\}$ using the language description of the subtask $l_n$. Then, we obtain policy candidates $\pi_{n}^k$ by fine-tuning the policy trained from the previous subtask $\pi_{n-1}^*$ with task code $(r_n^k, \rho_{g,n}^k)$.
    
    While previous works focused on evaluation metrics and feedback~\cite{ma2024eureka, song2023self} for evaluating policies with a LLM, we propose several strategies to identify the appropriate task code only with language description and without sample inefficient iterative searches. First, as proposed in~\ref{subsec:sequence of language}, our subtask descriptions are grounded in state variables. This approach minimizes ambiguity in translating natural language into task code. Moreover, we provide normalized state variables for defining the reward functions instead of raw state value. This enables the LLM to find proper weighting parameters in reward function generation without iterative trial-and-error. In our experiments, we found that no more than $5$ samples are required for capturing feasible task code among potential code error or variations in weight choices.

   It is crucial for the generated reward function to not only facilitate learning of the current subtask but also to \emph{remember} and utilize \emph{previously} learned subtasks. To achieve this, we record the history of trained subtasks and their corresponding task codes, denoted as $h$, and provide this information to the LLM when generating the task code for next subtask. This approach is analogous to the Chain-of-Thoughts prompting technique~\cite{wei2022chain}, enabling the generation of task code for complex subtasks in the later stages by building upon the simpler reward functions that were developed for earlier subtasks. Moreover, we instruct the LLM on how to adjust the weights of reward arguments: placing a greater emphasis on rewards related to the current task while still maintaining relevance to previous tasks to mitigate forgetting. This \emph{chain of task code} ensures a comprehensive and adaptive learning process across the curriculum.

\subsection{Large Language Model for Policy Evaluation}
\label{subsec: evaluation}

    After training $K$ different candidate policies $\pi_{n}^k,  k \in \{1,\dots, K\}$ using $K$ different task codes $(r_n^k, \rho_{g,n}^k), k \in \{1,\dots, K\}$, we evaluate the candidate polices and select the optimal policy $\pi_n^*$ that best aligns with current subtasks. While previous approaches relied on human feedback~\cite{liang2024learning} or predefined evaluation metrics~\cite{ma2024eureka, song2023self} to assess training outcomes, we evaluate the policy only with the language description of the current subtask. 
    
    To achieve this, we focused on trajectory analysis capabilities of LLMs shown in~\cite{wang2023prompt}. We provide the LLM with the statistics of trajectory rollout $\tau_n^1, \tau_n^2, \dots, \tau_n^K$ from each trained policy $\pi_n^1, \pi_n^2, \dots, \pi_n^K$. Similar to the task code generation process, task history information $h$ is given to the evaluation LLM. The LLM is then instructed to find the most effective task code for the current subtask description $l_n$, while also considering previous subtasks to avoid forgetting. This evaluation feedback mitigate randomness in LLM generated task code while ensuring learning with historical awareness.

    Finally, once the best policy $\pi_n^*$ for current subtask $m_n$ is identified, this policy is archived in the library and utilized as a pre-trained policy for subsequent subtask. Additionally, we update the history $h$ with the selected task code to keep track of the trained subtasks. Our overall algorithm is summarized in Algorithm~\ref{alg:main}.

    \begin{algorithm}
    \caption{Overall algorithm of CurricuLLM} \label{alg:main}
    \hspace*{\algorithmicindent} \textbf{Input} Language description of target task $l_T$, Language description of environment $l_E$ \\
    \hspace*{\algorithmicindent} \textbf{Hyperparameter} Number of task code samples $K$, Total number of training timesteps for each subtask $M$ 
    \begin{algorithmic}[1]
    \State \textcolor{mylinkcolor}{\# Generate curriculum from LLM} 
    \State $[l_1, l_2, \dots, l_N] \sim LLM_{Curriculum}(l_T, l_E)$  
    \State $\pi \sim \text{random policy}$ \quad \textcolor{mylinkcolor}{\# Initialize with random policy}
    \State $h = [\:]$ 
    \For{$l_n$ for $n=1:N$}
        \For{$k = 1:K$}
            \State \textcolor{mylinkcolor}{\# Sample task code from LLM} 
            \State $(r_{n}^k, \rho_{g,n}^k) \sim LLM_{Task Code Generation}(l_n, l_E, h)$
            \State $\pi_n^k \sim \text{train}(\pi, r_{n}^k, \rho_{g,n}^k, M)$
        \EndFor
        \State \textcolor{mylinkcolor}{\# Roll out trajectories from policy candidates}
        \State $\tau_n^1, \tau_n^2, \dots, \tau_n^K \sim \pi_{n}^1, \pi_n^2, \dots, \pi_n^K$ 
        \State \textcolor{mylinkcolor}{\# Evaluate the resulting trajectories} 
        \State $(r_n^*,\rho_{g,n}^*), \pi_n^* \sim LLM_{Evaluation}(l_n, \tau_n^1, \tau_n^2, \dots, \tau_n^K)$ 
        \State \textcolor{mylinkcolor}{\# Update the policy and training history } 
        \State $\pi \leftarrow \pi_n^*$ 
        \State $h.append(l_n, r_n^*, \rho_{g,n}^*)$ \quad 
    \EndFor
    \end{algorithmic}
    \hspace*{\algorithmicindent} \textbf{Output} Policy $\pi$
    \end{algorithm}

\section{Experiments}

\begin{figure}
    \centering
    \begin{subfigure}[b]{0.11\textwidth}
        \includegraphics[width=\textwidth, height=2cm]{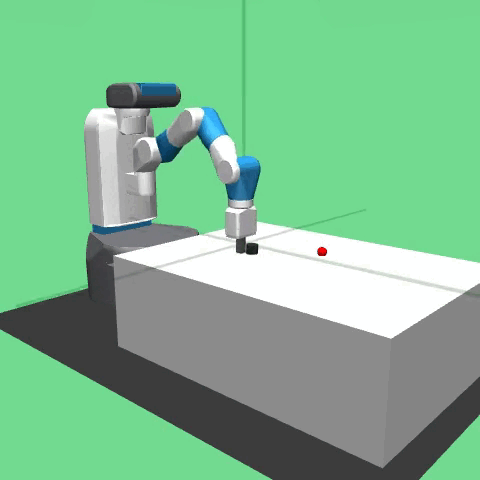}
        \label{fig:fetch_slide_env}   
    \end{subfigure}
    \begin{subfigure}[b]{0.11\textwidth}
        \includegraphics[width=\textwidth, height=2cm]{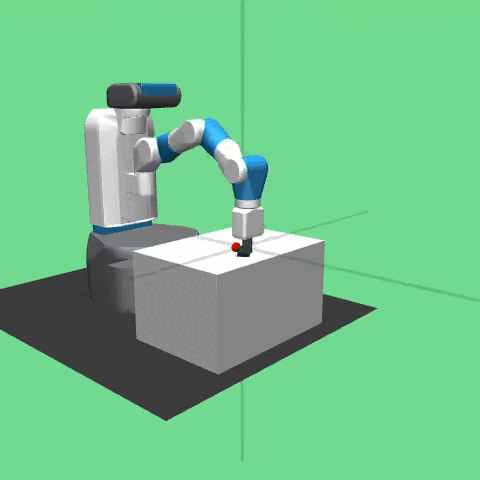}
        \label{fig:fetch_push_env}
    \end{subfigure}
    \begin{subfigure}[b]{0.11\textwidth}
        \includegraphics[width=\textwidth, height=2cm]{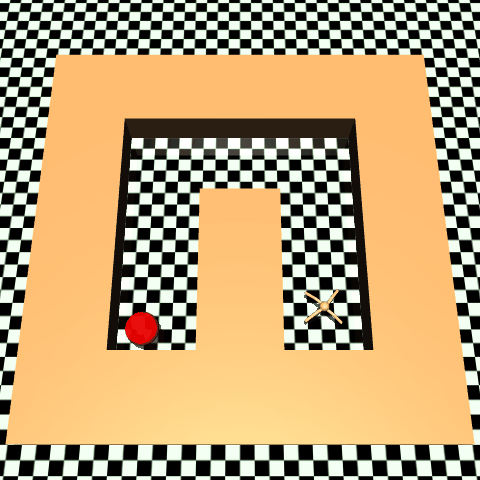}
        \label{fig:antmaze_env}
    \end{subfigure}
    \begin{subfigure}[b]{0.11\textwidth}
        \includegraphics[width=\textwidth, height=2cm]{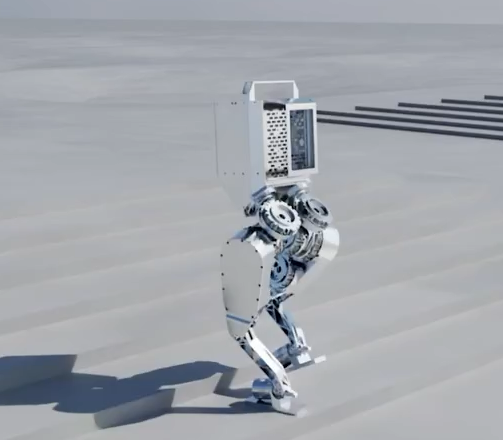}
        \label{fig:humanoid_env}
    \end{subfigure}
    \label{fig:env}
    \caption{Snapshot of Environments: From left to right, Fetch-Slide, Fetch-Push, AntMaze-UMaze, and Berkeley Humanoid.}
\end{figure}

\begin{figure*}
    \centering
    \begin{subfigure}[b]{0.3\textwidth}
        \includegraphics[width=\textwidth]{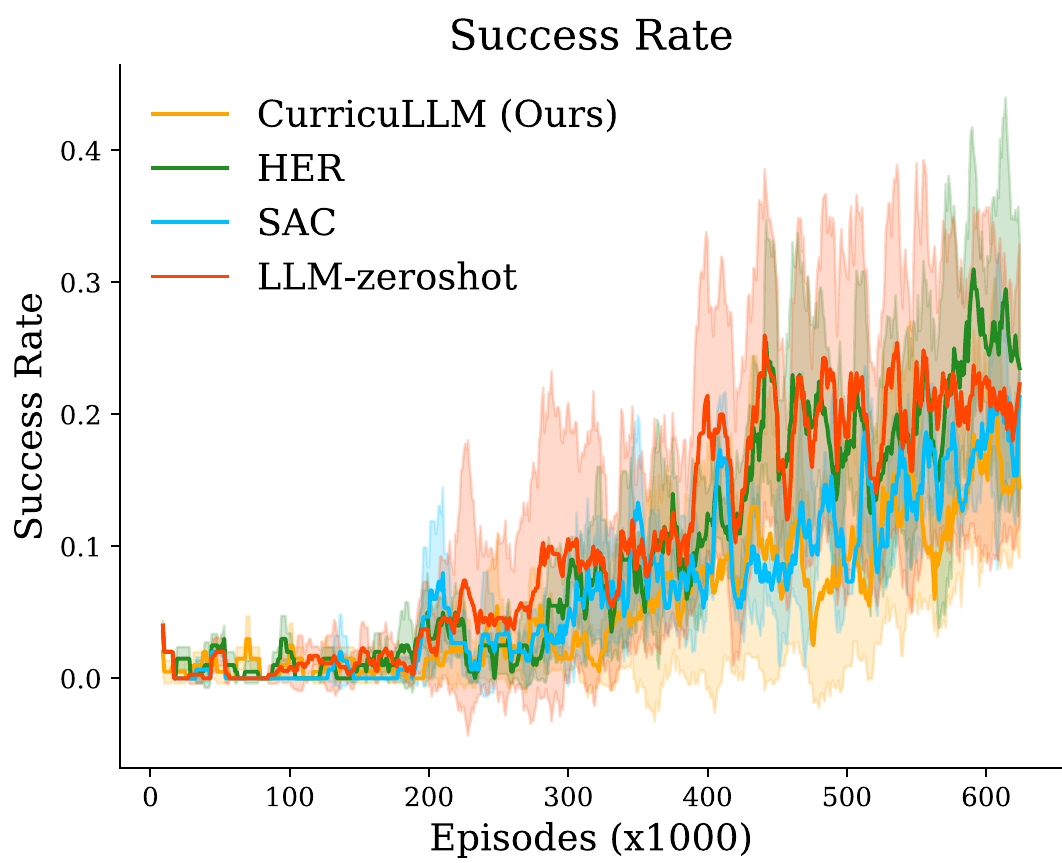}
        \caption{Fetch-Slide environment}
        \label{fig:fetch_slide}   
    \end{subfigure}
    \begin{subfigure}[b]{0.3\textwidth}
        \includegraphics[width=\textwidth]{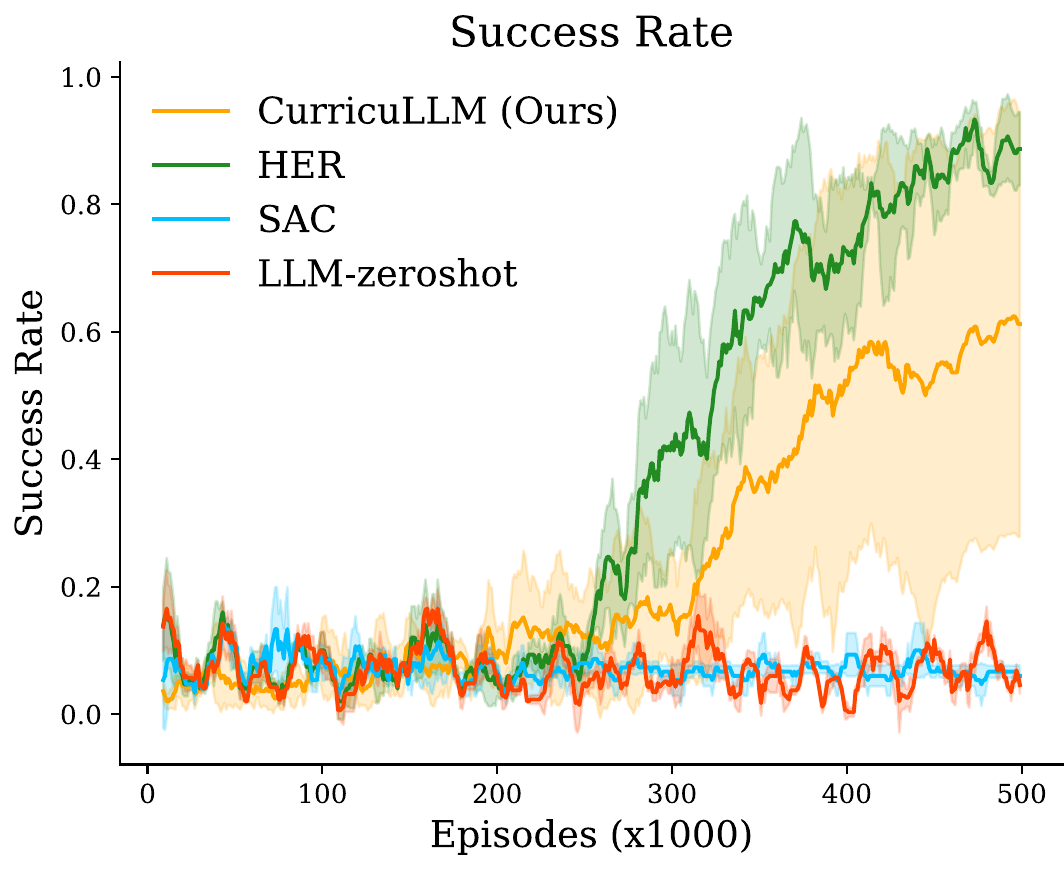}
        \caption{Fetch-Push environment}
        \label{fig:fetch_push}
    \end{subfigure}
    \begin{subfigure}[b]{0.3\textwidth}
        \includegraphics[width=\textwidth]{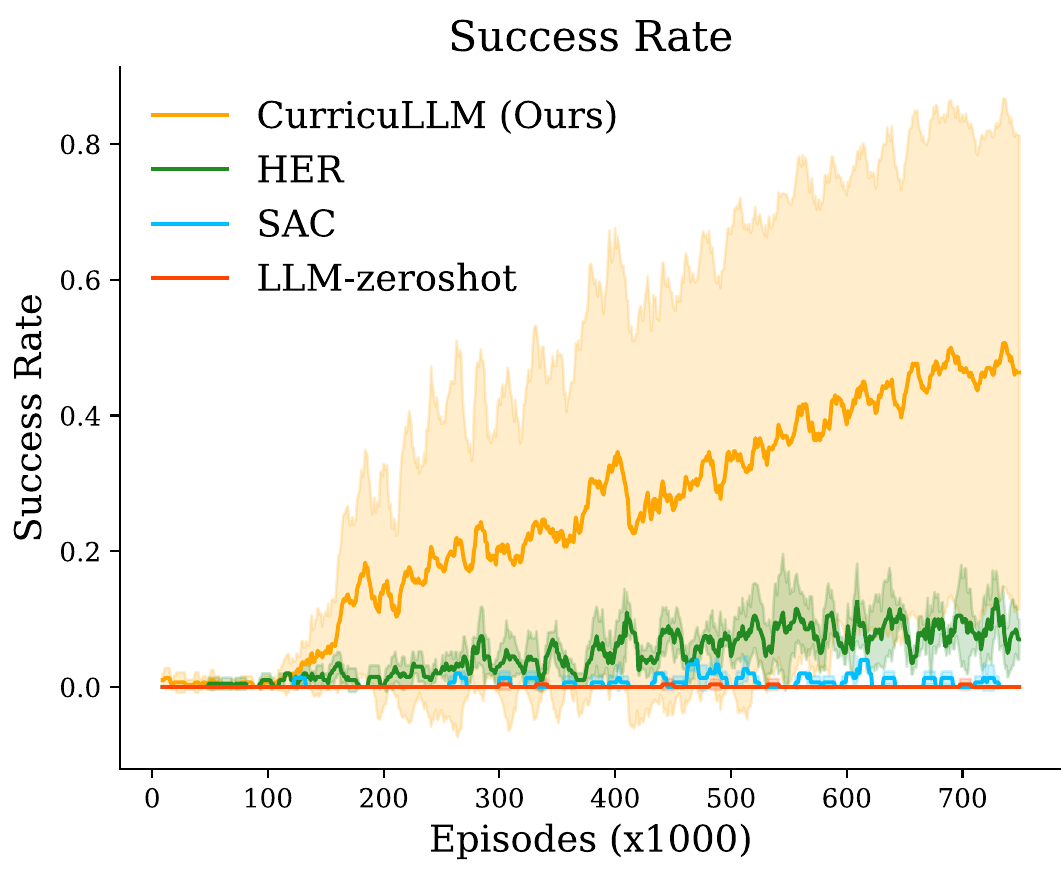}
        \caption{AntMaze-UMaze environment}
        \label{fig:antmaze}
    \end{subfigure}
    \caption{\small{Success rate of tasks in Gynmasium-Robotics environments. }}
    \label{fig:result}
\end{figure*}

In this Section, we evaluate CurricuLLM across different robotics task, ranging from manipulation, navigation, and locomotion, which covers broad range of robotic tasks. We first show CurricuLLM's efficiency in manipulation and navigation task in Section~\ref{subsec:gym}. Then, we evaluate CurricuLLM in higher dimension humanoid control problem in Section~\ref{subsec:humanoid}. Finally, we deploy the policy learned through CurricuLLM to Berkeley Humanoid~\cite{liao2024berkeley} to validate in hardware.

\subsection{Gymnasium Environments}
\label{subsec:gym}

For the manipulation and navigation tasks, we evaluate CurricuLLM in Fetch-Slide, Fetch-Push, and AntMaze-UMaze environments in Gymnasium-Robotics~\cite{gymnasium_robotics2023github} with Stable-Baselines3~\cite{stable-baselines3}. In the Fetch-Slide and Fetch-Push tasks, the 7DOF Fetch robot tries to manipulate a block to a desired goal position either by sliding or pushing the block. In the AntMaze environment, the ant robot tries to navigate the U-shaped maze and reach the goal position. While CurricuLLM was not able to change goal distributions in Fetch environments, in AntMaze, CurricuLLM was able to change the maximum distance between the initial ant position and the goal position in each subtasks. We compare our method with three baselines: soft actor-critic (SAC)~\cite{haarnoja2018soft}, hindsight experienced replay (HER)~\cite{andrychowicz2017hindsight}, and LLM-zeroshot. HER relabels previous uninformative trajectories as successful experience that achieved virtual goals within the trajectory, and then trains relabeled trajectories with off-policy RL algorithm. LLM-zeroshot baseline generates reward functions for the target task in a zero-shot manner from the LLM. Comparison with LLM-zeroshot highlights that CurricuLLM is necessary for learning complex control task and generating high-quality reward functions from LLM. While CurricuLLM is compatible with any RL algorithm, we implement with SAC to have a fair comparison with HER, which has to use an off-policy algorithm. Each algorithm was trained with $5$ different random seeds, while LLM-zeroshot is trained with $10$ different seeds to reflect randomness of LLM output. Finally, we use OpenAI's GPT-4-turbo~\cite{achiam2023gpt} as LLM agents throughout the experiments.


\noindent \textbf{Results.} The learning curves of the success rates for each target task are illustrated in Figure~\ref{fig:result}. In every task, especially in the Fetch-Push and AntMaze-UMaze, CurricuLLM exhibited comparable or superior success rate compared to the baselines. It is noteworthy that CurricuLLM outperformed every baseline in the most challenging AntMaze environment, which highlights the significance of curriculum in complex tasks. Furthermore, while HER is only applicable to goal-conditioned policy due to their relabeling scheme, CurricuLLM achieved comparable results in Fetch environments by only changing reward function throughout the curriculum. Finally, we mention that success rate of CurricuLLM was highly dependent on curriculum quality, resulting in a high variance in success rates. 

\begin{figure}
    \centering
    \includegraphics[width=0.8\columnwidth]{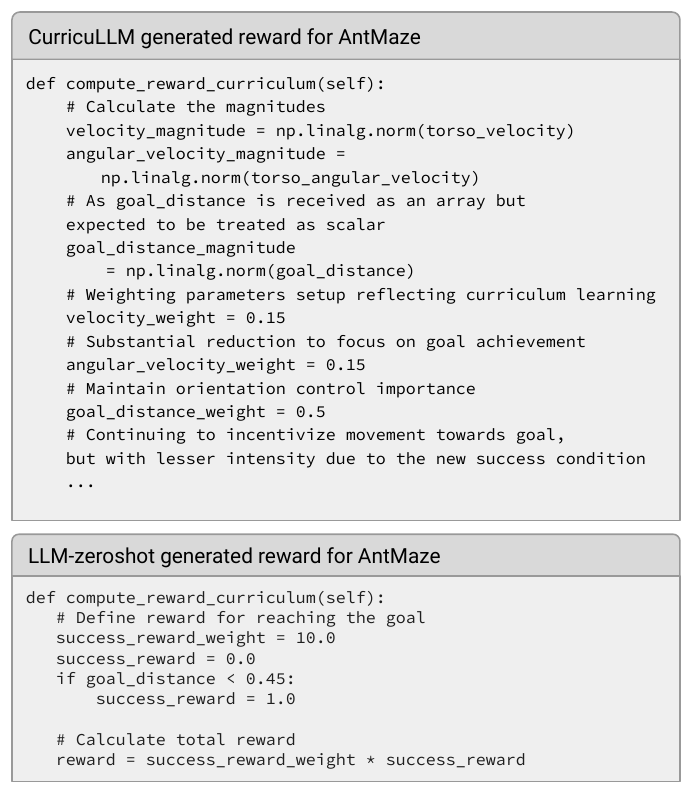}
    \caption{Example reward codes generated from CurricuLLM and reward code generated in a zero-shot manner from LLM for AntMaze environment. The reward code designed from CurricuLLM encompasses diverse behaviors that motivates learning the target task while directly querying a reward code for the target task leads to sparse-reward, uninformative reward function.}
    \label{fig:reward code}
\end{figure}

\noindent \textbf{Quality of Reward Code.} In Figure~\ref{fig:reward code}, we compare the final reward code generated by CurricuLLM with the reward code derived from LLM-zeroshot in AntMaze environment. We observe that the reward code from CurricuLLM captures diverse behaviors from subtasks which are beneficial for achieving the target task, such as incentivizing fast movement for exploration and regulating angular velocity for stabilization. In contrast, the reward code from LLM-zeroshot focuses solely on the target task without such terms that facilitate the learning. This demonstrates that our two-step approach—(1) generating a task-level curriculum and (2) evolving the task code effectively facilitates the generation of a informative reward function.

\subsection{Berkeley Humanoid}
\label{subsec:humanoid}

\subsubsection{Simulation Results}

To show the effectiveness of CurricuLLM in the higher dimension, complex locomotion task, we use Berkeley Humanoid~\cite{liao2024berkeley} environment implemented in Isaac Lab~\cite{mittal2023orbit}. Berkeley Humanoid is mid-size humanoid robot consist of state dimension of $45$ and action dimension of $12$, which is desired joint positions. The target task is making the robot run by following a random command within $\{(v_x, v_y, \theta)| -2 \leq v_x, v_y \leq 2, -\pi \leq \theta \leq \pi \}$, where $v_x, v_y$ are linear velocity in $x$ and $y$ direction, and $\theta$ is the heading angle. In curricula design, CurricuLLM was capable of specifying the command range and reward function for each subtasks. We compare with the policies trained with two human-designed reward functions, the one proposed in the original paper and the other adding alive bonus to such reward. We also compare with LLM-zeroshot method. We excluded HER since it is only applicable to goal-conditioned sparse reward environment. All algorithms are implemented with proximal policy optimization (PPO)~\cite{schulman2017proximal} as it is known for showing stable performance in locomotion tasks~\cite{li2024reinforcement, rodriguez2021deepwalk}. 

To ensure safe and desirable motion in real-world, we add pre-defined reward functions in addition to the LLM generated rewards, such as removing high-frequency actions or avoiding sliding motion. In addition, we add a termination condition based on the accumulated tracking error following~\cite{li2024reinforcement}. This termination condition is necessary for prohibiting the robot from achieving suboptimal behavior, such as sacrificing command tracking for stability.

\begin{table}
    \centering
    \caption{Results of Berkeley Humanoid Simulation}
    \begin{tabular}{|m{2cm}|m{1.6cm}|m{1.6cm}|m{1.7cm}|} 
        \hline
        & \makecell[l]{Tracking \\ Error (m/s)} & Failure (\%) & Episode Length (steps) \\
        \hline
        CurricuLLM & $\mathbf{0.46 \pm 0.38}$ & $\mathbf{8.79 \pm 4.72}$ & \makecell[l]{$315.44 \pm$ \\ $73.95$}  \\
        \hline
        LLM-zeroshot & $1.13 \pm 0.57$ & $28.57 \pm 45.19$ & \makecell[l]{$146.66 \pm$ \\ $114.12$} \\
        \hline
        Human reward & $\mathbf{0.41 \pm 0.10}$ & $\mathbf{7.31 \pm 3.12}$ & \makecell[l]{$397.67 \pm$ \\ $42.04$} \\
        \hline
        \makecell[l]{Human reward \\with live bonus} & $1.37 \pm 0.66$ & $0.51 \pm 0.23$ & \makecell[l]{$167.57 \pm$ \\ $98.22$} \\
        \hline
    \end{tabular}
    \label{table:humanoid}
\end{table}

\noindent \textbf{Results.} Simulation results are presented in Table~\ref{table:humanoid}. An episode is defined as failed if the robot terminated before 2 seconds. CurricuLLM achieved comparable results with the complex human designed reward in terms of tracking error and failure rate. Considering that human reward is designed after tedious trial-and-error from robot designer, it is significant that CurricuLLM achieved comparable results without any previous knowledge. LLM-zeroshot method tends to converge either to an unstable policy that terminates in the early stages or to a policy that sacrifices tracking error for stability. This is reflected in high the standard deviations of each performance metric. Interestingly, human reward with alive bonus converged to standing behavior that ignores the command. We suspect adding alive bonus made the policy to prioritize learning a stable policy over risking instability from robot movements. This illustrates that reward tuning is highly sensitive and requires extensive trial-and-error even for human experts, which motivates the automatic reward generation and training framework as CurricuLLM.

\subsubsection{Hardware Validation}

\begin{figure}
    \centering
    \includegraphics[width=0.9\linewidth]{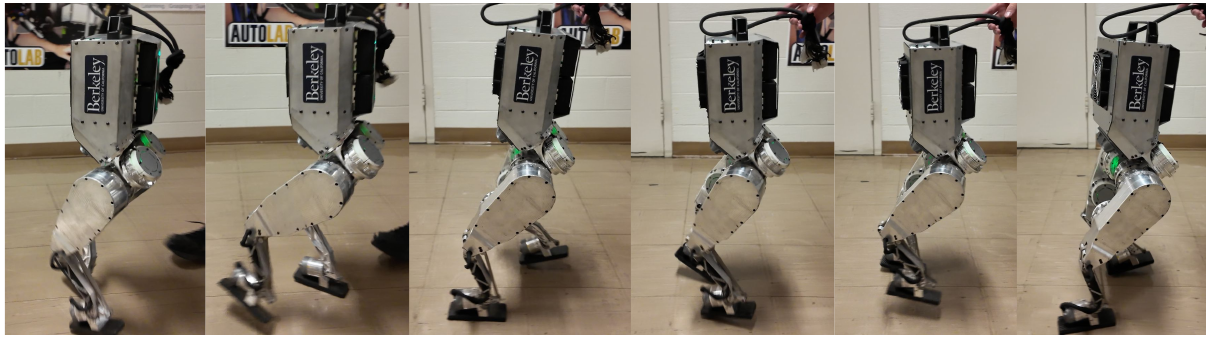}
    \caption{Snapshot of hardware experiment.}
    \label{fig:real-world snapshot}
\end{figure}

\begin{figure}
    \centering
    \includegraphics[width=0.9\linewidth]{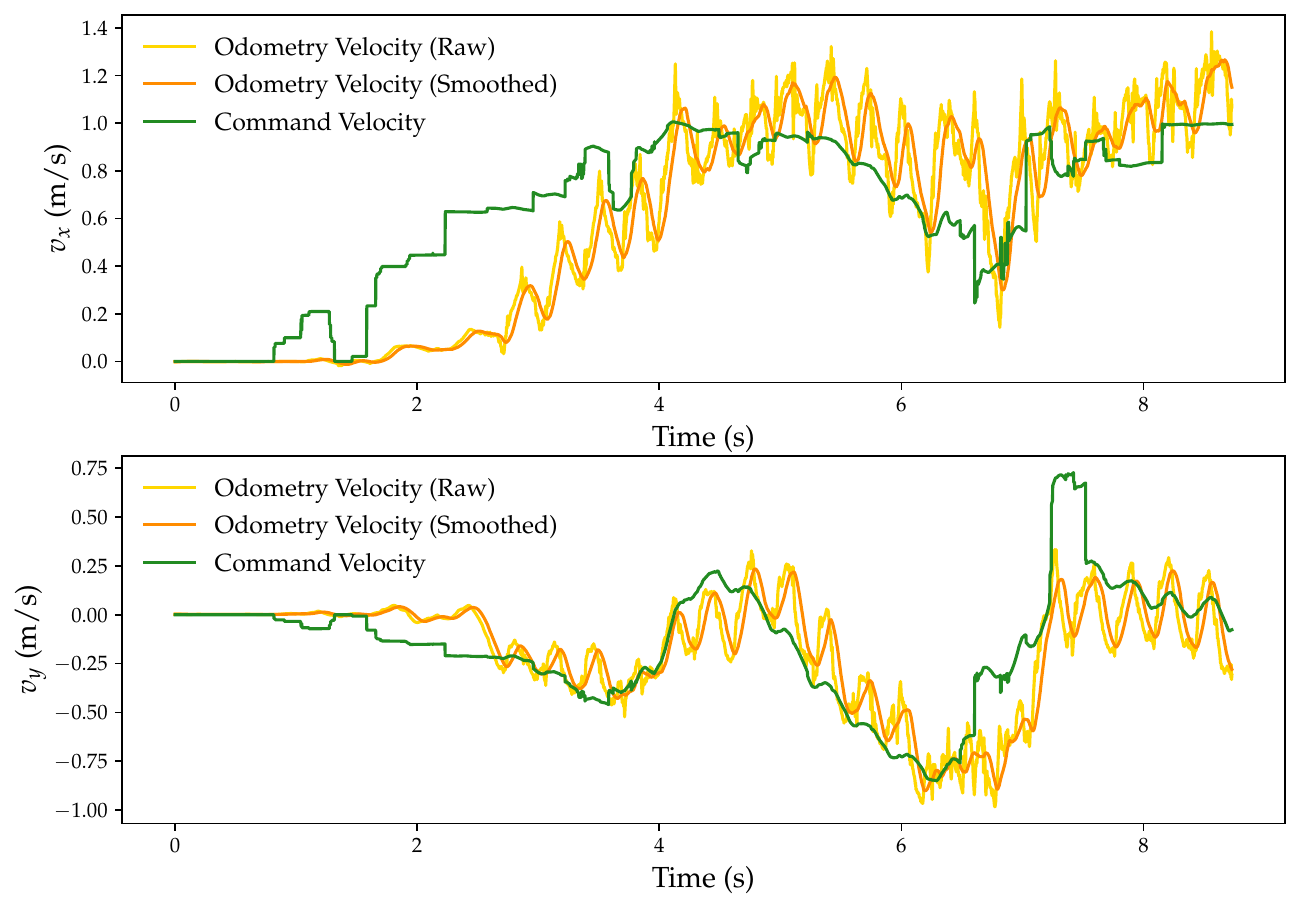}
    \caption{Odometry data and command data of base linear velocity $v_x$ in hardware experiments. The policy trained with CurricuLLM successfully follows the command in real-world.}
    \label{fig:hardware_experiments}
\end{figure}

In this section, we validate the locomotion policy learned through CurricuLLM on the Berkeley Humanoid. The command tracking performance in real-world experiment is given in Figure~\ref{fig:hardware_experiments}. In the hardware experiments, we had to restrict the demonstration to walking within the command velocity $\{(v_x, v_y, \theta)| -1 \leq v_x, v_y \leq 1, -\pi \leq \theta \leq \pi \}$ due to the restrictions that are specific to the hardware and not related to the control policy. Through our hardware experiments, we can found that the base velocity of the humanoid was able to accurately track the given command. This highlighted the effectiveness of the control policy learned through CurricuLLM in controlling the robot in real world.

\section{CONCLUSIONS}

In this paper, we introduced CurricuLLM, an automated task curriculum generator using LLMs. CurricuLLM first generate a language description of a sequence of subtasks that facilitate the learning of a complex target task. In each subtasks, the task code generation LLM translates the subtask descriptions into task codes, which includes a reward function and a goal distribution. After training policies for the sampled task codes, the evaluation LLM analyzes the trajectory of each policy and selects the best aligned policy for the subtask description. This optimal policy is utilized as a pre-trained model for subsequent subtasks. We showed that CurricuLLM successfully created task curricula and learned complex robotic control tasks in manipulation, navigation, and locomotion. We plan to extend our feedback loop by integrating training progress, such as learning curves, to optimize the training procedure by enabling adaptive training such as early stopping.









\bibliographystyle{ieeetr}
\bibliography{literature}

\newpage

\onecolumn
\section{APPENDIX}

\subsection{Prompts}
In this section, we provide example prompts for CurricuLLM. We present the prompt that we used for curriculum generation (Step 1), task code generation (Step 2), and policy evaluation (Step 3).

\renewcommand{\lstlistingname}{Prompt}
\begin{lstlisting}[language=, caption=Curriculum generation prompt for Berkeley Humanoid.]
You are a curriculum generator trying to generate a curriculum to solve reinforcement learning tasks as effectively as possible. 
Your goal is to write a list of tasks for the agent that will help the agent learn the main task in the environment described in the text.

Some helpful tips for generating a curriculum:
(1) Do not explore or learn dynamics by doing random actions. Each task should have a specific goal to achieve.
(2) Curriculum should be made as short as possible. Be concise.
(3) Never generate more than 5 tasks
(4) The last task of your curriculum should align with the original task description
(5) Focus on stability and learning healthy in early stages. However, focus on command tracking in later stages.

You will be given several variables you can use to describe each task in curriculum.
Each task should be described using the given variables and do not introduce new variables.
You should also describe the command that will be given to the robot.

You will be given several variables you can use to describe each task in curriculum.
Each task should be described using the given variables and do not introduce new variables.

Your output should be
Task 1
Name: []
Description: []
Reason: []

Task 2 
Name: []
Description: []
Reason: []

Task n
Name: Original task
Description: []
Reason: []

Fill out the inside of brackets in the format.
Remove the brackets after filling out them.

The hub is a bipedal robot that can walk and run in a fashin similar to that of humans or animals.
The original task in the environment is for hub to walk or run by following random command.
Speed of the hub robot will be specified in a command consist of [linear velocity x, linear velocity y, heading angle]
The command range in original task should follow this: "linear velocity x: [-2, 2] , linear velocity y: [-2, 2], heading angle [-pi, pi]"

In each subtask, you can train the agent for randomized command with specific command range.
You have to specify the range of command value in the task description.

Note that the default initial values are initialized to make hub stand.

The variables you can use to describe the curriculum includes

base_lin_vel: Linear velocity of base in xyz direction
base_ang_vel: angular velocity of base in xyz direction,
nonflat_base_orientation: Size of deviation of base orientation from default standing position, 
base_height_diff: height difference between measured base height and default stainding height, 
torques: Torques applied to the joint,
hip_pos_deviation: Relative position of hip joint from default standing position,
knee_pos_deviation: Relative position of knee joint from default standing position, 
joint_vel: Velocity of each joint, 
joint_acc: Acceleration of each joint, 
is_alive: 1 if the environment is healthy, 0 if the environment is terminated due to failure, 

commands: Range of commands that robot base should follow. 
          Consist of [linear velocity x range [], linear velocity y range [], heading angle range []].
\end{lstlisting}

\begin{lstlisting}[language=, caption=Task code generation prompt for Berkeley Humanoid.]
You should write reward function for given task using useful variables from the environment
Your reward function is part of the curriculum learning consist of learning sequence of different tasks.
You will be given description of past tasks, reward code for past tasks, and current task descriptiion.

Your reward_curriculum() function's return should be torch.Tensor of shape (number of environments)
The code output of reward_curriculum() should be formatted as a python code string: "```python ... ```".
Your have to specify your command range.
The code output should be formatted as "```command ... ```"

Some helpful tips for writing the reward function code:
(1) Use torch functions to write the reward function. Do not import additional library.
(2) If you want to get a size of value in 2d tensor, use torch.sum(torch.square(variable), dim=1)
(3) If you want to get a size of value in 1d tensor, use torch.square(variable)
(4) You may normalize each reward arguments
(5) You should use a weighting parameter outside of the transformation function
(6) You may put higher weight on the reward for current task. However, you should also give reward for past tasks to avoid forgetting.
(7) Focus on stability and learning healthy in early stages. However, focus on command tracking in later stages.
(8) Most importantly, you must only use given variables
(9) Do not change other predefined parts in the code.

The hub is a bipedal robot that can walk and run in a fashin similar to that of humans or animals.
The original task in the environment is for hub to walk or run by following random command.
Speed of the hub robot will be specified in a command consist of [linear velocity x, linear velocity y, angular velocity z, heading angle]
The command range in original task should follow this: "linear velocity x: [-2, 2] , linear velocity y: [-2, 2], heading angle is [-pi, pi]"

observation from normalized_obs() includes
base_lin_vel: Linear velocity of base in xyz direction, 2 dimension with shape (number of envs, 3)
base_ang_vel: angular velocity of base in xyz direction, 2 dimension with shape (number of envs, 3)
nonflat_base_orientation: Size of deviation of base orientation from default standing position, 1 dimension with shape (number of envs)
base_height_diff: height difference between measured base height and default stainding height, 1 dimension with shape (number of envs)
torques: Torques applied to the joint, 2 dimension with shape (number of envs, number of joints)
hip_pos_deviation: Relative position of hip joint from default standing position, 2 dimension with shape (number of envs, number of hip joints)
knee_pos_deviation: Relative position of knee joint from default standing position, 2 dimension with shape (number of envs, number of knee joints)
joint_vel: Velocity of each joint, 2 dimension with shape (number of envs, number of joints)
joint_acc: Acceleration of each joint, 2 dimension with shape (number of envs, number of joints)
is_alive: 1 if the environment is healthy, 0 if the environment is terminated due to failure, 1 dimension with shape (number of envs)

commands: Commands that robot base should follow. 
          Consist of [linear velocity x, linear velocity y, angular velocity z].
          2 dimension with shape (number ov envs, 3)

Note that default values are initialized to make the bipedal robot stand.

Your output for the environment code should follow this format
```python
def reward_curriculum(env: ManagerBasedRLEnv, asset_cfg: SceneEntityCfg = SceneEntityCfg("robot")) -> torch.Tensor:
    """Compute the reward for the current curriculum step."""
    command, base_lin_vel, base_ang_vel, hip_pos_deviation, knee_pos_deviation, \
    joint_vel, joint_acc, torque, \
    nonflat_base_orientation, base_height_diff = normalized_obs(env, asset_cfg)
    is_alive = (~env.termination_manager.terminated).float()

    n_envs = env.scene.num_envs

    # Implement your reward function here
    reward = torch.zeros(n_envs, device=env.device)
    return reward
```

You should re-implement reward_curriculum function to assign a proper reward function for given task. 
Note that you should try to follow command linear velocity and angular velocity z.
Especially, third component of base angular velocity should accurately follow angular velocity z in command.


Your output for the command value should follow this format.
You should change the range of lin_vel_x, lin_vel_y, and heading for your specific task.
You cannot change the range of ang_vel_z.
```command        
ranges=mdp.UniformVelocityCommandCfg.Ranges(lin_vel_x=(-2.0, 2.0), lin_vel_y=(-2.0, 2.0), ang_vel_z=(-1.0, 1.0), heading=(-math.pi, math.pi)),
```

Generate a reward function code and command for
Task Name: <<Task_Name>>
Description: <<Task_Description>>
Reason: <<Task_Reason>>

Previous learnd task was
Task Name: <<Task_Name>>
Description: <<Task_Description>>
Reason: <<Task_Reason>>
Code:
```python
<<Task_Code>>
```
\end{lstlisting}

\begin{lstlisting}[language=, caption=Policy evaluation prompt for Berkeley Humanoid]
You are a robotics engineer trying to evaluate the robot trajectory is following the given task.
You will be given a language Description of a task, and trajectory statistics of robots.
Based on the trajectory rollout information, choose the best agent for following the task Description.

Your task is part of curriculum learning consist of sequence of tasks.
While you should focus on evaluating the agent based on current task, also consider their performance on previous tasks.

Your priority is selecting healthy policy.
If episode length is shorter than 100, do not select it.

Follow the output format.

Decision: Agent [number]
Reason: 

Find the best agent for following task description in below

Task Name: <<Task_Name>>
Description: <<Task_Description>>
Reason: <<Task_Reason>>

Previous tasks in curriculum are
Task Name: <<Task_Name>>
Description: <<Task_Description>>
Reason: <<Task_Reason>>
\end{lstlisting}

\subsection{CurricuLLM Examples}
This section presents example curriculum, task codes, and evaluation decisions from CurricuLLM in Berkeley Humanoid.

\renewcommand{\lstlistingname}{Output}
\setcounter{lstlisting}{0}
\begin{lstlisting}[language=, caption=Example curriculum for Berkeley Humanoid]
Task 1
Name: Basic Locomotion
Description: In this task, commands will have a limited range to encourage initial movements without requiring precise control over speed or direction. The commands range should be linear velocity x: [-0.5, 0.5], linear velocity y: [-0.5, 0.5], heading angle: [-pi/4, pi/4].
Reason: To introduce the hub to basic movements and to ensure it learns to balance and take steps forward, backward, and to the sides within a controlled environment. This will also help stabilize its base and familiarize it with simple locomotion commands.

Task 2
Name: Advanced Locomotion
Description: Commands will be moderately challenging with increased range to enhance the hub's ability to move and change directions. The commands range should be linear velocity x: [-1, 1], linear velocity y: [-1, 1], heading angle: [-pi/2, pi/2].
Reason: To improve the hub's response to a wider range of motions and directions, teaching it to adapt its balance and coordination for faster and more varied movements. This serves as a bridge to mastering complex locomotion required in the original task.

Task 3
Name: Full Speed and Agility Training
Description: Commands will cover the full range specified for the original task to maximize the hub's movement capabilities. The commands range should be linear velocity x: [-2, 2], linear velocity y: [-2, 2], heading angle: [-pi, pi].
Reason: To ensure the hub is capable of handling the full spectrum of speed and directional changes, preparing it for any possible command within the original task's requirements. This task aims at fine-tuning the hub's motor control, agility, and responsiveness.

Task 4
Name: Original task
Description: Hub to walk or run by following random commands within the specified range: linear velocity x: [-2, 2], linear velocity y: [-2, 2], heading angle: [-pi, pi].
Reason: To assess the hub's ability to follow any given command accurately and efficiently, combining all learned skills from previous tasks. This final task evaluates how well the hub can integrate its training to achieve the goal of dynamic, responsive locomotion.
\end{lstlisting}

\renewcommand{\lstlistingname}{Code}
\setcounter{lstlisting}{0}
\begin{lstlisting}[language=python, caption=Example task code for Berkeley Humanoid. Task code for Basic Locomotion task.]
def reward_curriculum(env: ManagerBasedRLEnv, asset_cfg: SceneEntityCfg = SceneEntityCfg("robot")) -> torch.Tensor:
    """Compute the reward for the Basic Locomotion task."""
    command, base_lin_vel, base_ang_vel, hip_pos_deviation, knee_pos_deviation, \
    joint_vel, joint_acc, torques, \
    nonflat_base_orientation, base_height_diff = normalized_obs(env, asset_cfg)
    is_alive = (~env.termination_manager.terminated).float()

    n_envs = env.scene.num_envs

    # Locomotion Rewards
    reward_lin_vel_x = -torch.abs(command[:, 0] - base_lin_vel[:, 0])
    reward_lin_vel_y = -torch.abs(command[:, 1] - base_lin_vel[:, 1])
    reward_ang_vel_z = -torch.abs(command[:, 2] - base_ang_vel[:, 2])

    # Stability and Efficiency Rewards
    reward_base_orientation = -torch.abs(nonflat_base_orientation)
    reward_base_height_diff = -torch.abs(base_height_diff)
    reward_torques = -torch.sum(torch.abs(torques), dim=1)

    # Weighting the rewards
    weight_lin_vel_x = 0.2
    weight_lin_vel_y = 0.2
    weight_ang_vel_z = 0.4
    weight_base_orientation = 0.1
    weight_base_height_diff = 0.05
    weight_torques = 0.05
    weight_alive = 1.0

    # Combining the rewards
    reward = weight_lin_vel_x * reward_lin_vel_x + weight_lin_vel_y * reward_lin_vel_y + \
             weight_ang_vel_z * reward_ang_vel_z + weight_base_orientation * reward_base_orientation + \
             weight_base_height_diff * reward_base_height_diff + weight_torques * reward_torques + \
             weight_alive * is_alive

    return reward

ranges=mdp.UniformVelocityCommandCfg.Ranges(lin_vel_x=(-0.5, 0.5), lin_vel_y=(-0.5, 0.5), ang_vel_z=(-1.0, 1.0), heading=(-math.pi/4, math.pi/4)),
\end{lstlisting}

\begin{lstlisting}[language=Python, caption=Example task code for Berkeley Humanoid. Task code for final task which is following command within 2m/s.]
def reward_curriculum(env: ManagerBasedRLEnv, asset_cfg: SceneEntityCfg = SceneEntityCfg("robot")) -> torch.Tensor:
    """Compute the reward for the Original task."""
    command, base_lin_vel, base_ang_vel, hip_pos_deviation, knee_pos_deviation, \
    joint_vel, joint_acc, torques, \
    nonflat_base_orientation, base_height_diff = normalized_obs(env, asset_cfg)
    is_alive = (~env.termination_manager.terminated).float()

    n_envs = env.scene.num_envs

    # Original Task Rewards (evaluating adherence to any given command accurately and efficiently)
    reward_lin_vel_x = -torch.abs(command[:, 0] - base_lin_vel[:, 0])
    reward_lin_vel_y = -torch.abs(command[:, 1] - base_lin_vel[:, 1])
    reward_ang_vel_z = -torch.abs(command[:, 2] - base_ang_vel[:, 2]) * 1.5  # emphasize on angular velocity

    # Stability and Agility Rewards
    reward_base_orientation = -torch.abs(nonflat_base_orientation) * 0.5
    reward_base_height_diff = -torch.abs(base_height_diff) * 0.5
    reward_torques = -torch.sum(torch.abs(torques), dim=1) * 0.1  # adjusted to account for necessary torques during dynamic movements

    # Combining the rewards with adjusted weights to reflect the integration of all learned skills
    # Maintained proportion but slightly adjusted for overall performance and efficiency
    weight_lin_vel_x = 0.2
    weight_lin_vel_y = 0.2
    weight_ang_vel_z = 0.4  # Further emphasized for the original task
    weight_base_orientation = 0.05
    weight_base_height_diff = 0.05
    weight_torques = 0.05  # Slight adjustment for embracing dynamic actions
    weight_alive = 1.5  # Balance between surviving and task accomplishment

    # Final Reward Combination reflecting the task's goal
    reward = weight_lin_vel_x * reward_lin_vel_x + weight_lin_vel_y * reward_lin_vel_y + \
             weight_ang_vel_z * reward_ang_vel_z + weight_base_orientation * reward_base_orientation + \
             weight_base_height_diff * reward_base_height_diff + weight_torques * reward_torques + \
             weight_alive * is_alive

    return reward

ranges=mdp.UniformVelocityCommandCfg.Ranges(lin_vel_x=(-2.0, 2.0), lin_vel_y=(-2.0, 2.0), ang_vel_z=(-1.0, 1.0), heading=(-math.pi, math.pi)),
\end{lstlisting}

\renewcommand{\lstlistingname}{Output}
\setcounter{lstlisting}{1}
\begin{lstlisting}[language=, caption=Example evaluation result for Basic Locomotion task in Berkeley Humanoid. Average value from trajectory rollout is provided.]
Find the best agent for following task description in below

Task Name: Basic Locomotion
Description: In this task, commands will have a limited range to encourage initial movements without requiring precise control over speed or direction. The commands range should be linear velocity x: [-0.5, 0.5], linear velocity y: [-0.5, 0.5], heading angle: [-pi/4, pi/4].
Reason: To introduce the hub to basic movements and to ensure it learns to balance and take steps forward, backward, and to the sides within a controlled environment. This will also help stabilize its base and familiarize it with simple locomotion commands.

Agent 0:
base_lin_vel: [-0.031  0.023 -0.013]
base_ang_vel: [0.004 0.033 0.003]
projected_gravity: [ 0.444 -0.197 -0.87 ]
velocity_command: [ 0.461 -0.165  0.109]
hip_pos: [-0.366 -0.175 -0.362  0.096 -0.38  -0.71 ]
kfe_pos: [0.077 0.378]
ffe_pos: [-0.144 -0.174]
faa_pos: [-0.058  0.004]
joint_vel: [-0.016  0.023  0.085 -0.037 -0.086 -0.076  0.061  0.052  0.078  0.101
  0.351 -0.23 ]
last_actions: [-9.600e-02 -7.650e-01 -7.860e-01  1.520e-01 -3.640e-01 -1.033e+00
 -2.810e-01  1.000e-03  6.983e+00  1.152e+00  1.985e+00 -1.913e+00]
episode_length: 1000.0

Agent 1:
base_lin_vel: [-0.099 -0.166  0.012]
base_ang_vel: [0.007 0.003 0.127]
projected_gravity: [-0.042 -0.058 -0.995]
velocity_command: [-0.113 -0.489  0.113]
hip_pos: [ 0.46  -0.245  0.427 -0.27   0.027 -0.129]
kfe_pos: [0.019 0.237]
ffe_pos: [-0.044 -0.117]
faa_pos: [-0.075 -0.123]
joint_vel: [ 0.038  0.008  0.093  0.032 -0.037 -0.02   0.057  0.074  0.012  0.039
  0.343  0.193]
last_actions: [ 1.104 -0.784  0.728 -0.352 -0.11  -0.41  -0.821 -0.711 -3.689  0.257
  0.051  0.38 ]
episode_length: 1000.0

Agent 2:
base_lin_vel: [ 0.079 -0.289  0.003]
base_ang_vel: [-0.007  0.035 -0.143]
projected_gravity: [ 0.051  0.021 -0.994]
velocity_command: [ 0.172 -0.428 -0.176]
hip_pos: [ 0.309  0.16   0.036 -0.004  0.029 -0.101]
kfe_pos: [0.051 0.252]
ffe_pos: [-0.098 -0.052]
faa_pos: [-0.043 -0.1  ]
joint_vel: [ 0.024  0.001  0.063  0.032 -0.02  -0.035  0.052  0.058  0.03   0.111
  0.232  0.26 ]
last_actions: [ 0.352  0.266  0.466  0.102  0.028 -0.264 -0.262 -0.835  2.257 -3.157
  0.269 -0.447]
episode_length: 1000.0

Agent 3:
base_lin_vel: [ 0.023 -0.131 -0.383]
base_ang_vel: [2.006 0.577 0.047]
projected_gravity: [ 0.023 -0.063 -0.995]
velocity_command: [0.002 0.017 0.005]
hip_pos: [-0.073 -0.068  0.025  0.19  -0.169 -0.13 ]
kfe_pos: [0.28  0.258]
ffe_pos: [ 0.02  -0.032]
faa_pos: [-0.034  0.052]
joint_vel: [-2.201 -2.652  0.502  6.555 -4.148 -2.601  6.557  5.742  0.477 -0.715
 -1.158  0.67 ]
last_actions: [-0.181 -2.088 -0.139  3.195 -1.09  -0.826  1.922  1.813  1.214 -1.612
 -0.212 -0.147]
episode_length: 4.354

Decision: Agent 1
Reason: Among the agents, Agent 1 demonstrates the most relevant and balanced performance for the "Basic Locomotion" task. The task specifies a need for movements within a certain range of linear velocities and heading angles. Agent 1's trajectory data shows velocity commands (-0.113, -0.489, 0.113) and base linear velocities (-0.099, -0.166, 0.012) that are reasonably within or close to the specified command ranges for the task (linear velocity x and y: [-0.5, 0.5], heading angle: [-pi/4, pi/4]). 

The actions of Agent 1 also indicate movements that are modest and controlled compared to the other agents, which is suitable for basic locomotion learning and aligns with the aim of familiarizing the hub with simple motions. Agent 1's episode length, like all but Agent 3's, meets the criterion of being longer than 100 which reflects a healthy policy that can sustain activity over an acceptable duration.

Agent 0 and Agent 2 also show potential, but their velocity commands and base linear velocities are less aligned with the task's specified ranges than those of Agent 1. Agent 3 is disqualified for having an episode length (4.354) far below the acceptable threshold of 100, indicating an unhealthy or unstable policy for this task despite its possibly relevant actions and velocities, which cannot be reliably evaluated given the short episode length.

In consideration of the constrained environment meant to introduce basic movements, Agent 1's performance data suggests a balance between obeying the velocity command constraints and exhibiting a level of control conducive to learning basic movements.
\end{lstlisting}

\end{document}